\documentclass{article}

\usepackage[final]{nips_2017}

\usepackage[utf8]{inputenc} %
\usepackage[T1]{fontenc}    %
\usepackage{hyperref}       %
\usepackage{url}            %
\usepackage{booktabs}       %
\usepackage{amsfonts}       %
\usepackage{nicefrac}       %
\usepackage{microtype}      %
\usepackage{bbm}

\usepackage{graphicx}
\usepackage{amsmath}
\usepackage{amssymb}

\usepackage[english]{babel}
\usepackage[autostyle]{csquotes} %
\MakeOuterQuote{"}

\usepackage{gensymb}

\usepackage{subfig}

\usepackage[ruled]{algorithm2e}

\usepackage{authblk}

\title{Continual Learning in Generative Adversarial Nets}

\author[1]{Ari Seff}
\author[1]{Alex Beatson}
\author[1]{Daniel Suo}
\author[2]{Han Liu}
\affil[1]{Department of Computer Science, Princeton University}
\affil[2]{Department of Operations Research and Financial Engineering, Princeton University}

\begin{document}

\maketitle

\begin{abstract}
Developments in deep generative models have allowed for tractable learning of high-dimensional data distributions. While the employed learning procedures typically assume that training data is drawn i.i.d. from the distribution of interest, it may be desirable to model distinct distributions which are observed sequentially, such as when different classes are encountered over time. Although conditional variations of deep generative models permit multiple distributions to be modeled by a single network in a disentangled fashion, they are susceptible to catastrophic forgetting when the distributions are encountered sequentially. In this paper, we adapt recent work in reducing catastrophic forgetting to the task of training generative adversarial networks on a sequence of distinct distributions, enabling continual generative modeling. %
\end{abstract}

\section{Introduction}
Deep generative models have gained widespread use as a tractable way to model high-dimensional data distributions.
Recently introduced frameworks, such as generative adversarial networks (GANs) \cite{goodfellow2014} and variational autoencoders (VAEs) \cite{kingma2013}, can map a noise distribution to the data space, producing realistic samples. Capturing these high-dimensional data distributions makes possible a variety of downstream tasks, e.g., semi-supervised learning, sampling, and reconstruction/denoising. %

These models are also amenable to conditional training, where a conditional input (e.g., class label, data from another modality) guides the sampling process \cite{mirza2014, sohn2015}. At test time, manually selected conditional inputs lead to samples from the corresponding conditional distribution. Assuming the distributions share some underlying structure, this vastly reduces the capacity required to model the distributions compared to modeling each conditional distribution in a separate network.

However, the standard training regime for deep generative models assumes that training data is drawn i.i.d. from the distribution of interest. For example, in the setting where the dataset contains multiple classes (e.g., MNIST digits), data representing every class is used concurrently during training. In many real-world scenarios, data may actually arrive sequentially or only be available for a short time period.

Notably, a number of potential applications of deep generative models encounter exactly this constraint. One reason for widespread interest in unsupervised learning is that vastly more unlabeled data is available than labeled data. In many cases, such as an agent exploring an environment and learning a visual representation via a camera feed or training a generative model of text from the Twitter firehose, it is unlikely to be practical to save all the data over the lifetime of the system. A (batched) online or streaming approach is preferable. 

Another area which has gained recent interest and which exhibits this constraint is private learning \cite{dldp}. Learning a differentially private deep classifier from many private datasets currently requires many rounds of communication with or queries to models trained locally on each dataset \cite{papernot2016semi, federated}. While private learning of deep generative models is still an open research problem, a differentially private deep generative model would allow generation of synthetic data and good feature representations for downstream tasks, without the need for repeated communication. Given this model, it would be desirable to efficiently update it from private data sources in sequence without having to access previous data sources again.

If we wish to update a trained generative model to capture a newly observed distribution, naively training the model solely on the new data will result in the previously learned distributions being forgotten (Figure \ref{fig:add-on-vanilla}). This is due to the general susceptibility of neural networks to catastrophic forgetting when trained on multiple tasks sequentially \cite{mccloskey1989}. Instead, the standard training regime requires that we train on new and old data simultaneously. This approach is not very scalable as it requires that all previously observed data be stored, or synthetic data representative of previous observations be regenerated, for each round of training. 

In this work, we propose a scalable approach to training generative adversarial nets, a prominent deep generative model, on multiple datasets sequentially, where training data for any previous dataset is assumed to be inaccessible. We leverage recent work with discriminative models where synaptic plasticity is reduced for network parameters determined to be critical for previously encountered tasks \cite{kirkpatrick2016, zenke2017}. Experimental evaluation of our approach demonstrates that GANs may be extended to settings where the observed data distribution changes over time, enabling continual learning of novel generative tasks.

\section{Deep generative models}

In generative modeling, we wish to learn some distribution $p_{\textrm{data}}(x)$, e.g., the distribution of natural images. In a deep generative model, $p_{\textrm{data}}(x)$ is approximated by a neural network. Some generative models may be structured to allow for explicit estimation of the likelihood by directly approximating $p_{\textrm{data}}(x)$ with a function $p_\theta (x)$. Such models include variational autoencoders \cite{kingma2013} and their variants. Other generative models only allow for simulation of the data for sampling via a trainable, one-directional mapping from a fixed prior on latent variables $p_z(z)$ to the generator's output distribution:
\begin{equation}
x = G_\theta(z), \quad z \sim p_z(z) 
\end{equation}
In this case, the induced density is $p_g (x) = p_g(x|z) p_z(z)$, where $p_g(x|z) = \mathbbm{1} \{G_\theta(z) = x\}$. Such models notably include generative adversarial nets \cite{goodfellow2014}. While in this work we present a method for continual learning with GANs, we believe that similar routes may be taken with other generative modeling frameworks.  %

\subsection{Generative adversarial nets}
The standard GAN framework pits a generator $G$ against a discriminator $D$. Given a random noise input $z$ drawn from some prior $p_z(z)$ (typically a uniform or Gaussian distribution), $G$ deterministically generates a sample (e.g., an image). The generator's goal is to fool the discriminator such that the discriminator cannot determine if the generated samples come from $p_g(x) = G_\theta(z), z \sim p(z)$, $G$'s current learned distribution, or $p_{\textrm{data}}(x)$, the true data distribution. $G$ and $D$ are trained according to the following two-player minimax objective from \cite{goodfellow2014}:

\begin{equation}
\min_G \max_D V(D,G) = \mathbb{E}_{x\sim{p_{\textrm{data}}}}[\log D(x)] + \mathbb{E}_{z\sim{p_z(z)}}[\log (1 - D(G(z)))]
\end{equation} %

\subsection{Conditional generative adversarial nets}
In \cite{mirza2014}, the GAN framework is extended to allow both $G$ and $D$ to be conditioned on an extra input $y$. The minimax objective then becomes:

\begin{equation} \label{eq:minimax}
\min_G \max_D V(D,G) = \mathbb{E}_{x\sim{p_{\textrm{data}}}}[\log D(x, y)] + \mathbb{E}_{z\sim{p_z(z)}}[\log (1 - D(G(z, y), y))]
\end{equation}

This conditional framework is quite flexible, both in terms of what the input $y$ can be and how it enters the generator and discriminator networks. For example, $y$ may range from being a discrete, one-hot vector representing a class label to being a partially complete sample, as in the case of image inpainting. Furthermore, $y$ may simply be included as input to the first layers of $D$ and $G$, or it may be input at every layer. This is considered an architectural hyperparameter, and certain ways of incorporating $y$ may be more effective than others depending on the specific problem domain.  %

\section{Catastrophic forgetting and remedies}

When a neural network is trained on multiple tasks sequentially, the process of training on the later tasks tends to decrease performance on the earlier tasks. This phenomenon, known as catastrophic forgetting, was first explored in \cite{mccloskey1989}. A classic experiment in that work consisted of first training a multilayer perceptron with backpropagation on "ones" addition tasks (e.g., $1+5$, $1+8$) until it learned to correctly compute the sums. Then, the network was trained on "twos" addition tasks, without further access to the training data or loss function for ones addition. As the network learned to perform twos addition, it forgot how to performs ones addition. 

This same phenomenon is observable across diverse domains in which neural networks are widely used, including computer vision, speech recognition, and natural language processing. When network weights can adjust without restriction to meet the objective of the current task, previous task performance is severely disrupted. 

To our knowledge, existing literature on preventing catastrophic forgetting in neural networks has focused on discriminative/classification settings. Some approaches aim to directly preserve the functional input/output mapping of the network for previous tasks. For example, in \cite{li2016}, when new task training data arrives, the softmax activations of previous task output layers are recorded for all new training examples. Then, in a similar process to knowledge distillation \cite{hinton2015}, these softmax activations serve as targets for the old output layers while training on the new task. Other approaches make dramatic architectural changes to reduce interference between different tasks. For example, in \cite{rusu2016}, distinct columns are added to a network for each task while freezing parameters in all previous columns. Lateral connections allow the sharing of features between tasks. While forgetting is prevented, the number of parameters grows quadratically in the number of tasks. In our work, we assume a setting where even linear growth in the number of parameters is undesirable.

\subsection{Protecting critical parameters}
Kirkpatrick et al. \cite{kirkpatrick2016} recently proposed an approach, dubbed elastic weight consolidation (EWC), that is computationally inexpensive and does not require adding any new capacity to the network. Those parameters found to be critical for the performance achieved on previous tasks are protected with an additional loss term restricting their movement. Given a network parameterized by $\theta$ trained on some task $A$ with training data $D_A$, the posterior distribution of the parameters, $p(\theta | D_A)$, is approximated as a Gaussian distribution with mean $\theta^*_A$ (the actual parameter values after training) and variance given by the inverse diagonal of the Fisher information matrix $F$ (or Hessian). Thus those parameters associated with higher Fisher diagonal elements are considered to be more "certain" and thus less flexible. When the network begins training on a new task $B$ without further access to task $A$'s training data, the loss function for task $B$ is augmented as:

\begin{equation}
L(\theta) = L_B(\theta) + \sum_i \frac{\lambda}{2} F_i (\theta_i - \theta^*_{A,i})^2
\end{equation}

where $L_B$ is the standard loss for task $B$, $F_i$ is the $i^{th}$ diagonal element of the Fisher information matrix, and $\lambda$ is a hyperparameter denoting the relative importance of the tasks. Experimental results in \cite{kirkpatrick2016} on permuted MNIST tasks and Atari games validated this approach for classification tasks in supervised and reinforcement learning settings.

A similar method is proposed in \cite{zenke2017}. Rather than compute the weight saliency at the end of a task's training session, an online measure takes into account a weight's total contribution to the global change in loss over a training trajectory. In this case, those weights with higher contributions are more critical. When moving to subsequent tasks, a quadratic penalty is used similar to \cite{kirkpatrick2016}.

These approaches have connections to \cite{lecun1990}, where the second derivative of the objective function with respect to the network parameters, in combination with parameter magnitude, is used to estimate saliency. Low saliency parameters are pruned, leading to reduced network size. Rather than completely removing low saliency parameters, \cite{kirkpatrick2016, zenke2017} allow such parameters to have greater flexibility when learning a new task.

\section{Continual learning with GANs}

Assume we have a generator trained to simulate some distribution of interest. When the observed data distribution changes, we may wish to additionally model this new distribution without interfering with the old. While we could certainly train an entirely new GAN, for similar distributions (e.g., images of cats and images of dogs), this approach prevents the leveraging of any shared underlying structure. In addition, storing separate models for each task is not scalable in limited capacity settings. %

One might try to simply continue training the model according to the standard objective, only using data from the new distribution. If the network trains according to this setup however, it quickly undergoes catastrophic forgetting (Figure \ref{fig:add-on-vanilla}). Without being exposed to real examples from the previously encountered data, $D$ learns to determine if its inputs are real examples sampled from the new distribution only. Therefore, all of $G$'s resources will be directed towards fooling $D$ solely on the new task, leading to an inability to sample from the previously learned distributions. %

A simple yet expensive solution to $G$'s forgetting problem is to regenerate a training set for each of $G$'s previously encountered datasets, a route unavailable in the discriminate learning setting. Assuming $G$ has been trained well, the samples it generates should be realistic enough to serve as "real" inputs to $D$. Then standard training can proceed simultaneously on all the data, both new and old.

While this route reduces the extent of catastrophic forgetting, it is not very efficient. For the $t\textsuperscript{th}$ observed distribution, we must regenerate training sets for the previous $t-1$ distributions. In addition, we must retrain on all of the regenerated data in addition to the newly observed data. Learning from a sequence of $t$ observed distributions would then have a time complexity of $O(t^2)$. Instead, the approach we ultimately take is $O(t)$, requiring no regeneration of training data. 

\subsection{Augmented generator objective}

We propose to replace $G$'s portion of the standard minimax objective with an augmented loss function discouraging those weights critical for accurately modeling the previously observed distribution(s) from undergoing drastic changes in value. However, in the unconditional GAN framework, this constraint would attempt to preserve the mapping from the single noise prior $p_z(z)$ to the previous data distribution, while $D$'s objective would encourage mapping the noise prior to the new data distribution. We observed that $D$ is able to overcome arbitrarily large fixed constraints, ultimately resulting in the forgetting of $G$'s previously learned distribution. %

In the conditional GAN framework, the distinct observed data distributions may be associated with distinct conditional inputs, maintaining a well-defined desired mapping from the joint distribution of $z$ and a conditional input $y$ to the data space. In settings where $p_{data}(x|y)$ does not change over time but $p_y(y)$ does, such as when datasets corresponding to each class are observed sequentially, we may directly apply the constraints to a conditional GAN. In settings where conditional inputs are not available or the observed distribution changes arbitrarily over time, we may artificially assign a distinct conditional input $y$ for the distributions observed at different times.%

The portion of the standard minimax objective for conditional GANs relevant to determining $G$'s parameter updates is:

\begin{equation}
\label{eq:g-ob}
\mathbb{E}_{z\sim{p_z(z)}, y\sim{p_y(y)}}[\log (1 - D(G(z,y), y))]
\end{equation}

In practice, we follow the standard approach of using $- \log (D(G(z,y), y))$ as the generator's cost function instead to provide better gradients early in training. 

When $G$ has been trained well on an initial set of conditional inputs leading to optimal parameters $\theta^*_A$ for task $A$, we do not want subsequent task training sessions to significantly damage the performance achieved. Intuitively, the distribution of $D(G(z,y), y)$ should not be greatly altered for task $A$. Estimating the Fisher information of $D(G(z,y), y)$ with respect to $G$'s parameters allows us to determine those that are most critical to fooling $D$, and we can correspondingly penalize the movement of these parameters. Constraining movement of parameters with high Fisher information for previous tasks was introduced for discriminative models in \cite{kirkpatrick2016} as elastic weight consolidation (EWC).  %

We define the penalty weighting for $G$'s $i\textsuperscript{th}$ parameter as: %

\begin{equation}
F_i = \mathbb{E}_{z\sim{p_z(z)}, y\sim{p_y(y)}}\Big[\big(\frac{\partial}{\partial \theta_i} \log (D(G(z,y),y))\big)^2 \big \vert \theta = \theta^*_A \Big]
\end{equation}

Those parameters that, if perturbed, would have a high effect on the discriminator's output have a high value for the corresponding element in the diagonal of the Fisher information matrix and are considered especially salient. Note that the above actually corresponds to an approximation known as the empirical Fisher information, which is more efficient to compute \cite{martens2016}.

Now, $G$'s objective is augmented as:

\begin{equation}
L(\theta) = \mathbb{E}_{z\sim{p_z(z)}, y\sim{p_y(y)}}[- \log D(G(z,y),y))] + \sum_i \frac{\lambda}{2} F_i (\theta_i - \theta^*_{A,i})^2
\end{equation}

 \cite{zenke2017} explores a method similar to EWC for constraining the movement of salient parameters. Unlike in EWC, the computation of parameter importance in \cite{zenke2017} is performed online and does not require the between-task phase for estimating Fisher information. This approach could alternatively be used here.

\section{Experimental Results}
We evaluate our approach with class-conditional GANs, in a setting where new image classes are observed over time. Here, the classes are digit categories from the MNIST \cite{mnist} and SVHN \cite{svhn} datasets, and the conditional input $y$ is a one-hot encoding of the class label.

Unlike the typical concurrent training regime of a class-conditional GAN, in the continual learning setup the distinct conditional distributions are encountered sequentially, where previously observed data is unavailable. While the proposed framework assumes that the same generator network is used to model each new distribution, it does not require that the same discriminator is used. Here, we experiment in the extreme setting where, similar to previously observed data, all previously trained discriminators are assumed inaccessible.

Architecturally, this setup may either be interpreted as new elements being appended to $y$ during each training session, with the corresponding initialization of new connections, or as some element of $y$ simply being always zero when the corresponding class is not part of the current task. We use the Adam optimization algorithm \cite{adam} and re-initialize the optimizer parameters for each task. Thus, no updates occur for weights with zero gradient, making the two implementations equivalent.

On MNIST, we test our approach with a simple multilayer perceptron (MLP) GAN, while on SVHN we use a deep convolutional GAN (DCGAN, \cite{radford2015}).

\subsection{MLP + MNIST}
For MNIST, we first employ a simple GAN architecture where both $D$ and $G$ are two-layer MLPs. The hidden layers of both networks are of size 128 with ReLU activation. In $G$, the output layer is of size 784 with sigmoid activation, corresponding to the generated image. $D$ has a single output with sigmoid activation, corresponding to the estimated probability that the input example was drawn from $p_\textrm{data}$.

\subsubsection{Overcoming forgetting}
\begin{figure}
\centering 
\includegraphics[width=0.8\textwidth]{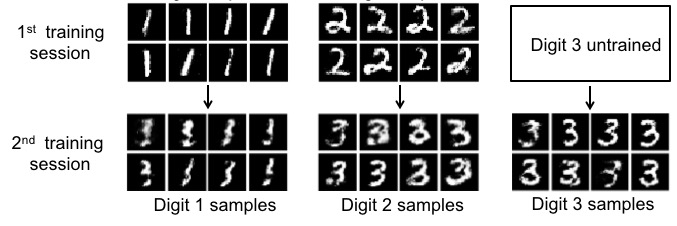}
\caption{Conditionally sampled images of the digit 1 and digit 2 after training with the standard objective (top) and sampled images of the digits 1, 2, and 3 after training a new conditional input for digit 3 (bottom). When using the standard conditional GAN objective, the generator forgets how to sample from the previously learned distributions.}
\label{fig:add-on-vanilla}
\end{figure}

\begin{figure}
\centering 
\includegraphics[width=0.8\textwidth]{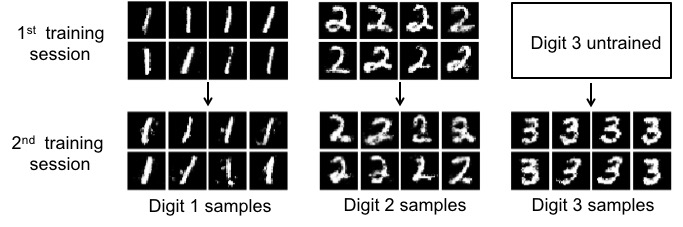}
\caption{Conditionally sampled images of the digit 1 and digit 2 after training with the standard objective (top) and sampled images of the digits 1, 2, and 3 after training a new conditional input for digit 3 using the EWC-augmented objective (bottom). The generator no longer forgets how to produce samples from the previous categories.}
\label{fig:add-on-ewc}
\end{figure}

\begin{figure}
\centering 
\includegraphics[width=\textwidth]{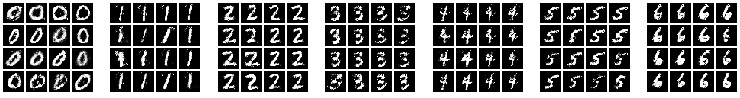}
\caption{Images sampled from a generator after training on different classes sequentially. Digits 0 and 1 were first trained concurrently, and the remaining digit classes were encountered one at a time sequentially. All samples were drawn after the last training session (after digit 6).}
\label{fig:multi-sequential}
\end{figure}

In Figure \ref{fig:add-on-ewc}, we see the samples produced by a generator first trained concurrently on the digits 1 and 2. When digit 3 is encountered in the next training session and the generator's objective is augmented with EWC, the previous distributions are not forgotten. This is in contrast to the catastrophic forgetting evident in Figure \ref{fig:add-on-vanilla}, where the generator trains on digit 3 with the standard objective. In Figure \ref{fig:multi-sequential}, we see samples from a generator first trained concurrently on digits 0 and 1, and then trained sequentially on the digits 2 through 6 with the augmented objective. Every displayed image was sampled after the final training session (on digit 6), and it is clear the previously learned distributions are not forgotten.

While catastrophic forgetting is significantly reduced when using the augmented objective, a small amount of peripheral noise is evident in the generated samples (e.g., in the digit 1 images in Figure \ref{fig:add-on-ewc}. In general, the discriminative capability of $D$ tends to be local to $p_g(x)$ and $p_\textrm{data}(x)$. That is, $D$ only learns to distinguish between real examples and the most recently generated samples. Images which are neither realistic nor similar to the current generated distribution are not meaningfully classified by $D$. 

\begin{figure}
\centering 
\includegraphics[width=0.55\textwidth]{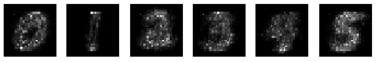}
\caption{Visualization of digit-specific Fisher information. A conditional GAN is trained on the digits $0$ through $5$ concurrently, and then the pixel-wise mean Fisher information for $G$'s output is computed per each conditional input.}
\label{fig:fish_visual}
\end{figure}

This is evident when visualizing the saliency of $G$'s output layer parameters (Figure \ref{fig:fish_visual}). The most critical parameters for fooling $D$ are those corresponding to the structure of the digit. Once $G$ matures enough during training such that peripheral noise is no longer produced, $D$ does not need to devote resources to detecting noise pixels. This limitation could potentially be addressed by training $D$ with experience replay, or by adding noise to some proportion of fake images $D$ is trained on. In practice, we find that higher capacity networks are less susceptible to this limitation.

\begin{figure}
\centering 
\includegraphics[width=\textwidth]{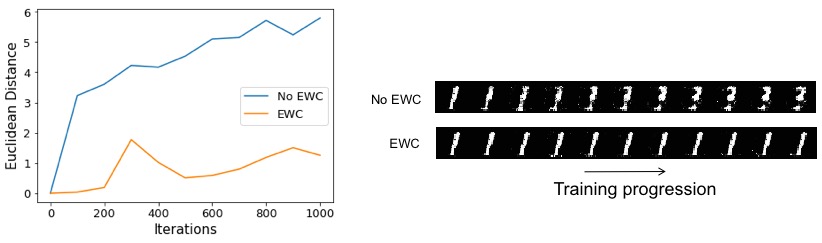}
\caption{Sampling from $G$ at a fixed $z$ while training a new conditional input. Images of the digit 1 are sampled at the same fixed $z$ while training on the digit 3 with the standard objective (top right) and with the EWC-augmented objective (bottom right). Catastrophic forgetting is only visible with the standard objective. The Euclidean distance of the current sampled image from the original image is shown on the left under both training objectives.} %
\label{fig:fixed_z} %
\end{figure}

Figure \ref{fig:fixed_z} shows the deformation to an image sampled from a previously learned distribution at a fixed $z$ during the first epoch of training on a new class, with and without $G$'s EWC-augmented objective. While training on the digit 3 with the standard objective, $G$ forgets how to produce visually coherent images of the digit 1. The digit 1 image transitions to a skinny 3 as training progresses. No deformation is apparent when the augmented objective is used. 


\subsection{DCGAN + SVHN}
For SVHN, we use a conditional DCGAN architecture with three convolutional layers and one fully connected layer in both $D$ and $G$. We append $y$ to $z$ in the input to $G$, while in $D$ a trainable linear transformation of $y$ is added to the output of the first convolutional layer.
\subsubsection{Overcoming forgetting}

\begin{figure}%
    \centering    
    \subfloat[]{{\includegraphics[width=2.965cm]{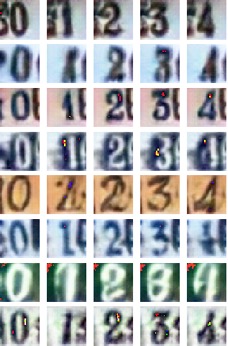} }}%
    \quad
    \subfloat[]{{\includegraphics[width=6cm]{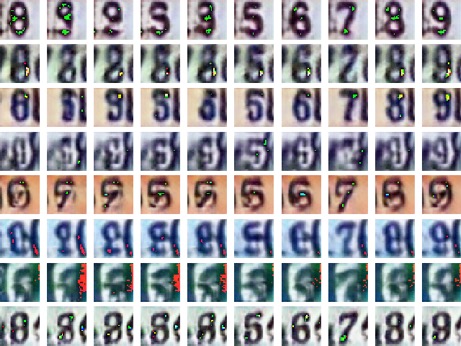} }}%
    
     \subfloat[]{{\includegraphics[width=6cm]{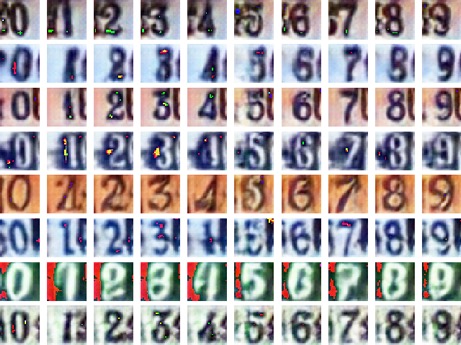} }}%
    
    \caption{Overcoming forgetting for the SVHN dataset. Each column represents a fixed $y$ while each row represents a fixed $z$. (a) Conditionally sampled images from a generator trained on the digits 0 through 4. (b) Images sampled from the same generator after continuing to train on the digits 5 through 9 with the standard objective. (c) Images sampled from the same generator after training with the EWC-augmented objective on the digits 5 through 9.}%
    \label{fig:svhn}%
\end{figure}

Figure \ref{fig:svhn} shows outputs from (a) a DCGAN model trained on digits zero to four concurrently, (b) the same model after continuing training with digits five to nine, and (c) the same model after continuing training with the augmented loss on digits five to nine. Forgetting is clearly present in (b). In (c), the visual quality of the digits zero to four is similar to the quality of those digits in (a), while the visual quality of the digits five to nine is similar to the quality of those digits in (b).

\subsection{Invariance to $\lambda$}
In both MNIST + MLP and SVHN + DCGAN, we observed that the results of applying EWC were largely invariant to the magnitude of $\lambda$. Our reported results for MNIST used $\lambda = 1000$, while reported results for SVHN used $\lambda = 100$. For SVHN we also trained models with $\lambda = 1$ and $\lambda = 5000$ and observed very little difference in visual quality and no difference in diversity in the final results. An argument could be made for $\lambda = 1$ being the most principled parameter when interpreting EWC as a Taylor expansion of the previous task's loss. However, with small values of $\lambda$ the generator parameters tended to rapidly deviate from their previous task values at the beginning of training on a new task, resulting in poor quality samples, before gradually being pulled back and visual quality increasing over the course of training. With high values of $\lambda$ we observed no loss in visual fidelity when beginning training on a new task, with rapid convergence to a model that produced good samples for all classes.

\section{Conclusion}
We have introduced an approach for continual learning with generative adversarial nets. Experimental results demonstrate that sequential training on different sets of conditional inputs utilizing an EWC-augmented loss counteracts catastrophic forgetting of previously learned distributions. The approach is general and applicable to any setting where the observed distribution of conditional inputs (e.g., class label, partially complete sample) changes over time, or where a conditional input representing the time of data capture can be appended to the data. 

The promise of deep generative modeling is the ability to learn a model of the world from vast sources of data; to fully realize its potential, it will be necessary to learn from data at the point of capture rather than storing all of it for future use. This work is one step towards this goal.

\subsubsection*{Acknowledgments}
This work was supported in part by the Department of Defense through the National Defense Science \& Engineering Graduate Fellowship (NDSEG) Program.

\bibliographystyle{plain}
\bibliography{references}

\end{document}